# Event-based Information Extraction for the biomedical domain: the Caderige project

**Erick Alphonse\*\*, Sophie Aubin\*, Philippe Bessières\*\*, Gilles Bisson\*\*\*\*, Thierry Hamon\*, Sandrine Lagarrigue\*\*\*, Adeline Nazarenko\*, Alain-Pierre Manine\*\*, Claire Nédellec\*\*, Mohamed Ould Abdel Vetah\*\*, Thierry Poibeau\*, Davy Weissenbacher\***

*\*Laboratoire d'Informatique de Paris-Nord*
*CNRS UMR 7030*
*Av. J.B. Clément 93430 F-Villetaneuse*
*{firstname.lastname}@lipn.univ-paris13.fr*

*\*\*\*Laboratoire de Génétique Animale,*
*INRA-ENSAR*
*Route de Saint Brieuc, 35042 Rennes Cedex*
*lagarrig@roazhon.inra.fr*

*\*\*Laboratoire Mathématique, Informatique et Génome (MIG),*
*INRA,*
*Domaine de Vilvert, 78352 F-Jouy-en-Josas*
*{firstname.lastname}@jouy.inra.fr*

*\*\*\*\*Laboratoire Leibniz – UMR CNRS 5522*
*46 Avenue Félix Viallet - 38031 F-Grenoble Cedex*
*Gilles.Bisson@imag.fr*

## Abstract

This paper gives an overview of the Caderige project. This project involves teams from different areas (biology, machine learning, natural language processing) in order to develop high-level analysis tools for extracting structured information from biological bibliographical databases, especially Medline. The paper gives an overview of the approach and compares it to the state of the art.

## 1 Introduction

Developments in biology and biomedicine are reported in large bibliographical databases either focused on a specific species (e.g. Flybase, specialized on *Drosophilia Menogaster*) or not (e.g. Medline). This type of information sources is crucial for biologists but there is a lack of tools to explore them and extract relevant information. While recent named entity recognition tools have gained a certain success on these domains, event-based Information Extraction (IE) is still a challenge. The Caderige project aims at designing and integrating Natural Language Processing (NLP) and Machine Learning (ML) techniques to explore, analyze and extract targeted information in biological textual databases. We promote a corpus-based approach focusing on text pre-analysis and normalization: it is intended to drain out the linguistic variation dimension, as most as possible. Actually, the MUC (1995) conferences have demonstrated that extraction is more efficient when performed on normalized texts. The extraction patterns are thus easier to acquire or learn, more abstract and easier to maintain

Beyond extraction patterns, it is also possible to acquire from the corpus, via ML methods, a part of the knowledge necessary for text normalization as shown here.

This paper gives an overview of current research activities and achievements of the Caderige project. The paper first presents our approach and compares it with the one developed in the framework of a similar project called Genia (Collier *et al.* 1999). We then propose an account of Caderige techniques on various filtering and normalization tasks, namely, sentence filtering, resolution of named entity synonymy, syntactic parsing, and ontology learning. Finally, we show how extraction patterns can be learned from normalized and annotated documents, all applied to biological texts.

## 2 Description of our approach

In this section, we give some details about the motivations and choices of implementation. We then briefly compare our approach with the one of the Genia project.

## 2.1 Project organization

The Caderige project is a multi disciplinary French research project on the automatic mining of textual data from the biomedical domain and is mainly exploratory orientated. It involved biology teams (INRA), computer science teams (LIPN, INRA and Leibniz-IMAG) and NLP teams (LIPN) as major partners, plus LRI and INRIA from 2000 to 2003.

## 2.2 Project motivations

Biologists can search bibliographic databases via the Internet, using keyword queries that retrieve a large superset of relevant papers. Alternatively, they can navigate through hyperlinks between genome databanks and referenced papers. To extract the requisite knowledge from the retrieved papers, they must identify the relevant abstracts or paragraphs. Such manual processing is time consuming and repetitive, because of the bibliography size, the relevant data sparseness, and the database continuous updating. From the Medline database, the focused query "Bacillus subtilis and transcription" which returned 2,209 abstracts in 2002, retrieves 2,693 of them today. We chose this example because *Bacillus subtilis* is a model bacterium and *transcription* is a central phenomenon in functional genomics involved in genic interaction, a popular IE problem.

```
GerE stimulates cotD transcription and
inhibits cotA transcription in vitro by
sigma K RNA polymerase, as expected from
in vivo studies, and, unexpectedly,
profoundly inhibits in vitro
transcription of the gene (sigK) that
encode sigma K.
```

**Figure 1:** A sentence describing a genic interaction

Once relevant abstracts have been retrieved, templates should be filled by hand since there is no available IE tool operational in genomics

| Interaction | **Type**: positive |
|---|---|
| | **Agent**: *GerE* |
| | **Target**: transcription of the gene sigK |

**Figure 2:** A template describing a genic interaction.

Still, applying IE *à la* MUC to genomics and more generally to biology is not an easy task because IE systems require deep analysis methods to locate relevant fragments. As shown in the example in Figures 1 and 2, retrieving that GerE is the agent of the inhibition of the transcription of the gene sigK requires at least syntactic dependency analysis and coordination processing. In most of the genomics IE tasks (function, localization, homology) the methods should then combine the semantic-conceptual analysis of text understanding methods with IE through pattern matching.

## 2.3 Comparison with the Genia project

Our approach is very close to the one of the Genia project (Collier *et al.*, 1999). Both projects rely on precise high-level linguistic analysis to be able to perform IE. The kind of information being searched is similar, concerning mainly gene and protein interaction as most of the research in this domain. The Genia corpus (Ohtae *et al.* 2001) is not specialized on a specific species whereas ours is based on *Bacillus Subtilis*.

Both projects develop annotation tools and Document Type Definition (DTD), which are, for the most part, compatible. The aim here is to build training corpus to which various techniques of NLP and ML are applied in order to acquire efficient event-based extraction patterns. The choice of ML and NLP methods differs but their aim is similar to our: normalizing text with predicate-arguments structures for learning better patterns. For example, Genia uses a combination of parsers to finally perform an HPSG-like analysis. The Caderige syntactic analysis is based on the specialization of the Link Parser (Sleator and Temperley, 1993 see section 4) to the biological domain.

In the following two sections, we detail our text filtering and normalization methods. Filtering aims at pruning the irrelevant part of the corpus while normalization aims at building an abstract representation of the relevant text. Section 4 is devoted to the acquisition of extraction patterns from the filtered and normalized text.

# 3 Text filtering

IR and text filtering are a prerequisite step to IE, as IE methods (including normalization and learning) cannot be applied to large and irrelevant corpora (they are not robust enough and they are computationally expensive). IR here is done through Medline interface by keyword queries for filtering the appropriate document subset. Then, text filtering, reduces the variability of textual data with the following assumptions:

- desired information is local to sentences ;
- relevant sentences contain at least two gene names.

These hypotheses may lead to miss some genic interactions, but we assume that information redundancy is such that at least one instance of each interaction is contained into a single sentence in the corpus. The documents retrieved are thus segmented into sentences and the sentences with at least two gene names are selected.

To identify the only relevant sentences among thoses, classical supervised ML methods have been applied to a *Bacillus Subtilis* corpus in which relevant and irrelevant sentences had been annotated by a biological expert. Among SVMs, Naïve Bayes (NB) methods, Neural Networks, decision trees (Marcotte *et al.*, 2001; Nedellec *et al.*, 2001), (Nedellec *et al,* 2001) demonstrates that simple NB methods coupled with feature selection seem to perform well by yielding around 85 % precision and recall. Moreover, our first experiments show that the linguistic-based representation changes such as the use of lemmatization, terminology and named entities, do not lead to significant improvements. The relevant sentences filtered at this step are then used as input of the next tasks, normalization and IE.

# 4 Normalization

This section briefly presents three text normalization tasks: normalization of entity names, normalization of relations between text elements through syntactic dependency parsing and semantic labeling. The normalization process, by providing an abstract representation of the sentences, allows the identification of regularities that simplify the acquisition or learning of pattern rules.

## 4.1 Entity names normalization

Named Entity recognition is a critical point in biological text analysis, and a lot of work was previously done to detect gene names in text (Proux and al., 1998), (Fukuda and al., 1998). So, in Caderige, we do not develop any original NE extraction tool. We focus on a less studied problem that is synonyms recognition.

Beyond typographical variations and abbreviations, biological entities often have several different names. Synonymy of gene names is a well-known problem, partly due to the huge amount of data manipulated (43.238 references registered in Flybase for Drosophilia Melanogaster for example). Genes are often given a temporary name by a biologist. This name is then changed according to information on the concerned gene: for example SYGP-ORF50 is a gene name temporarily attributed by a sequencing project to the PMD1 yeast gene. We have shown that, in addition to available data in genomic database (GenBank, SwissProt,…), it is possible to acquire many synonymy relations with good precision through text analysis. By focusing on synonymy trigger phrases such as "also called" or "formerly", we can extract text fragments of that type : `gene trigger gene.`

However, the triggers themselves are subject to variation and the arguments of the synonymy relation must be precisely identified. We have shown that it is possible to define patterns to recognize synonymy expressions. These patterns have been trained on a representative set of sentences from Medline and then tested on a new corpus made of 106 sentences containing the keyword *formerly*. Results on the test corpus are the following: 97.5% precision, 75% recall. We chose to have a high precision since the acquired information must be valid for further acquisition steps (Weissenbacher, 2004).

The approach that has been developed is very modular since abstract patterns like `gene trigger gene` (the trigger being a linguistic marker or a simple punctuation) can be instantiated by various linguistic items. A score can be computed for each instantiation of the pattern, during a learning phase on a large representative corpus. The use of a reduced tagged corpus and of a large untagged corpus justify the use of semi-supervised learning techniques.

## 4.2 Sentence parsing

The extraction of structured information from texts requires precise sentence parsing tools that exhibit relevant relation between domain entities. Contrary to (Akane *et al.* 2001), we chose a partial parsing approach: the analysis is focused on relevant parts of texts and, from these chunks, on specific relations. Several reasons motivate this choice: among others, the fact that relevant information generally appears in predefined syntactic patterns and, moreover,

| Rel | nbRel | Link Parser | | | | HCP | | | |
|---|---|---|---|---|---|---|---|---|---|
| | | relOK | R. | RelTot | P. | RelOK | R | RelTot | P. |
| **Subject** | 18 | 13 | 0.72 | 19 | 0.68 | 14 | 0.78 | 20 | 0.65 |
| **Object** | 18 | 16 | 0.89 | 17 | 0.94 | 9 | 0.5 | 13 | 0.69 |
| **Prep** | 48 | 25 | 0.52 | 55 | 0.45 | 20 | 0.42 | 49 | 0.41 |
| **V-GP1** | 14 | 13 | 0.93 | 15 | 0.87 | 9 | 0.64 | 23 | 0.39 |
| **O-GP** | 16 | 7 | 0.43 | 12 | 0.58 | 12 | 0.75 | 28 | 0.43 |
| **NofN** | 16 | 13 | 0.81 | 15 | 0.87 | 14 | 0.87 | 26 | 0.54 |
| **VtoV** | 10 | 9 | 0.9 | 9 | 1 | 7 | 0.7 | 7 | 1 |
| **VcooV** | 10 | 8 | 0.8 | 9 | 0.89 | 6 | 0.6 | 6 | 1 |
| **NcooN** | 10 | 8 | 0.7 | 10 | 0.8 | 4 | 0.4 | 6 | 0.67 |
| **nV-Adj** | 10 | 8 | 0.8 | 9 | 0.89 | 0 | 0 | 0 | 1 |
| **PaSim** | 18 | 17 | 0.94 | 18 | 0.94 | 17 | 0.94 | 22 | 0.77 |
| **PaRel** | 12 | 11 | 0.92 | 11 | 1 | 8 | 0.67 | 11 | 0.73 |

**Table 1:** Evaluation of two parsers on various syntactic relations
Relations meaning: subject = subject-verb, Object = verb-object, Prep = prepositional phrase, V-GP = verb-prep. phrase, O-GP = Object- prep. phrase, NofN = Noun of noun, VtoV = Verb to Verb, VcooV = Verb coord. Verb, NcooN = Noun coord. Noun, nV-Adj = not + Verb or adjective, PaSim = passive form, PaRel = passive relative the fact that we want to learn domain knowledge ontologies from specific syntactic relations (Faure and Nedellec, 2000 ; Bisson *et al.* 2000).

First experiments have been done on several shallow parsers. It appeared that constituent based parsers are efficient to segment the text in syntactic phrases but fail to extract relevant functional relationships betweens phrases. Dependency grammars are more adequate since they try to establish links between heads of syntactic phrases. In addition, as described in Schneider (1998), dependency grammars are looser on word order, which is an advantage when working on a domain specific language.

Two dependency-based syntactic parsers have been tested (Aubin 2003): a hybrid commercial parser (henceforth HCP) that combines constituent and dependency analysis, and a pure dependency analyzer: the Link Parser.

Prasad and Sarkar (2000) promote a twofold evaluation for parsers: on the one hand the use of a representative corpus and, on the other hand, the use of specific manually elaborated sentences. The idea is to evaluate analyzers on real data (corpus evaluation) and then to check the performance on specific syntactic phenomena. In this experiment, we chose to have only one corpus, made of sentences selected from the Medline corpus depending on their syntactic particularity. This strategy ensures representative results on real data.

A set of syntactic relations was then selected and manually evaluated. This led to the results presented for major relations only in table 1. For each analyzer and relation, we compute a recall and precision score (recall = # relevant found relations / # relations to be found;

precision = # relevant found relations / # relations found by the system).

The Link Parser generally obtains better results than HCP. One reason is that a major particularity of our corpus (Medline abstracts) is that sentences are often (very) long (27 words on average) and contain several clauses. The dependency analyzer is more accurate to identify relevant relationships between headwords whereas the constituent parser is lost in the sentence complexity. We finally opted for the Link Parser. Another advantage of the Link Parser is the possibility to modify its set of rules (see next subsection). The Link parser is currently used in INRA to extract syntactic relationships from texts in order to learn domain ontologies on the basis of a distributional analysis (Harris 1951, Faure and Nédellec, 1999).

### 4.3 Recycling a general parser for biology

During the evaluation tests, we noticed that some changes had to be applied either to the parser or to the text itself to improve the syntactic analysis of our biomedical corpus. The corpus needs to be preprocessed: sentence segmentation, named entities and terms recognition are thus performed using generic modules tuned for the biology domain[1]. Term recognition allows the removing of numerous structure ambiguities, which clearly benefits the parsing quality and execution time.

---

[1] A term analyser is currently being built at LIPN using existing term resources like Gene Ontology (see Hamon and Aubin, 2004).

Concerning the Link Parser, we have manually introduced new rules and lexicon to allow the parsing of syntactic structures specific to the domain. For instance, the Latin-derived Noun Adjective phrase "Bacillus subtilis" has a structure inverse to the canonical English noun phrase (Adjective Noun). Another major task was to loosen the rules constraints because Medline abstracts are written by biologists who express themselves in sometimes broken English. A typical error is the omission of the determinant before some nouns that require one. We finally added words unknown to the original parser.

## 4.4 Semantic labelling

Asium software is used to semi-automatically acquire relevant semantic categories by distributional semantic analysis of parsed corpus. These categories contribute to text normalization at two levels, disambiguating syntactic parsing and typing entities and actions for IE. Asium is based on an original ascendant hierarchical clustering method that builds a hierarchy of semantic classes from the syntactic dependencies parsed in the training corpus. Manual validation is required in order to distinguish between different meanings expressed by identical syntactic structures.

# 5 Extraction pattern learning

Extraction pattern learning requires a training corpus from which the relevant and discriminant regularities can be automatically identified. This relies on two processes: text normalization that is domain-oriented but not task-oriented (as described in previous sections), and task-oriented annotation by the expert of the task.

## 5.1 Annotation procedure

The Caderige annotation language is based on XML and a specific DTD (Document Type Definition that can be used to annotate both *prokaryote* and *eukaryote* organisms by 50 tags with up to 8 attributes. Such a precision is required for learning feasibility and extraction efficiency. Practically, each annotation aims at highlighting the set of words in the sentence describing:

- Agents (A): the entities activating or controlling the interaction

- Targets (T): the entities that are produced or controlled
- Interaction (I): the kind of control performed during the interaction
- Confidence (C): the confidence level in this interaction.

The annotation of "A low level of GerE activated transcription of CotD by GerE RNA polymerase in vitro ..." is given below. The attributes associated to the tag `<GENIC-INTERACTION>` express the fact that the interaction is a transcriptional activation and that it is certain. The other tags (`<IF>`, `<AF1>`, …) mark the agent (AF1 and AF2), the target (TF1) and the interaction (IF).

```
<GENIC-INTERACTION
     id="1"
     type="transcriptional"
     assertion="exist"
     regulation="activate"
     uncertainty="certain"
     self-contained="yes"
     text-clarity="good">
<IF>A<I> low level </I>of</IF>
<AF1><A1
        type=protein
        role=modulate
        direct=yes> GerE
</A1></AF1>,
<IF><I>activated</I> transcription
     of</IF>
<TF1><T1 type=protein> CotD </T1>
     </TF1> by
<AF2><A2
        type=protein
        role=required>
     GerE RNA polymerase
</A2></AF2>,
<CF>but<C>in vitro</C></CF>
</GENIC-INTERACTION>
```

## 5.2 The annotation editor[2]

Annotations cannot be processed in text form by biologists. The annotation framework developed by Caderige provide a general XML editor with a graphic interface for creating, checking and revising annotated documents. For instance, it displays the text with graphic attributes as defined in the editor XML style sheet, it allows to add the tags without strong constraint on the insertion order and it automatically performs some checking.

The editor interface is composed of four main parts (see Figure 3). The editable *text* zone for annotation, the list of XML *tags* that can be used at a given time, the *attributes* zone to edit the values of the selected tag, and the *XML*

---

[2] Contact one of the authors if you are interested to use this annotation tool in a research project

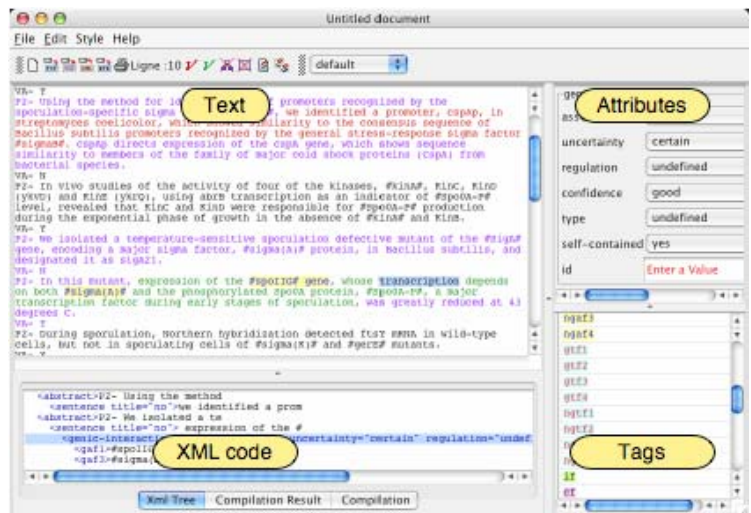

**Figure 3**: the Caderige annotation editor

*code* currently generated. In the text zone, the above sentence is displayed as follows:

> A low level of GerE activated transcription of CotD by GerE RNA polymerase but in vitro

This editor is currently used by some of the Caderige project partners and at SIB (Swiss Institute of BioInformatics) with another DTD, in the framework of the European BioMint project. Several corpora on various species have been annotated using this tool, mainly by biologists from INRA.

## 5.3 Learning

The vast majority of approaches relies on hand-written pattern rules that are based on shallow representations of the sentences (e.g. Ono et al., 2001). In Caderige, the deep analysis methods increase the complexity of the sentence representation, and thus of the IE patterns. ML techniques appear therefore very appealing to automate the process of rule acquisition (Freitag, 1998; Califf et al., 1998; Craven et al., 1999).

Learning IE rules is seen as a discrimination task, where the concept to learn is a n-ary relation between arguments which correspond to the template fields. For example, the template in figure 2 can be filled by learning a ternary relation genic-interaction(X,Y,Z), where X,Y and Z are the type, the agent and the target of the interaction. The learning algorithm is provided with a set of positive and negative examples built from the sentences annotated and normalized. We use the relational learning algorithm, Propal (Alphonse et al., 2000). The appeal of using a relational method for this task is that it can naturally represent the relational structure of the syntactic dependencies in the normalized sentences and the background knowledge if needed, such as for instance semantic relations.

For instance, the IE rules learned by Propal extract, from the following sentence :"In this mutant, expression of the spoIIG gene, whose transcription depends on both sigA and the phosphorylated Spo0A protein, Spo0AP, a major transcription factor during early stages of sporulation, was greatly reduced at 43 degrees C.", successfully extract the two relations genic-interaction(positive, sigA, spoIIG) and genic-interaction(positive, Spo0AP, spoIIG). As preliminary experiments, we selected a subset of sentences as learning dataset, similar to this one. The performance of the learner evaluated by ten-fold cross-validation is $69\pm6.5\%$ of recall and $86\pm3.2\%$ of precision. This result is encouraging, showing that the normalization process provides a good representation for learning IE rules with both high recall and high precision.

## 6 Conclusion

We have presented in this paper some results from the Caderige project. Two major issues are the development of a specific annotation editor for domain specialists and a set of machine learning and linguistic processing tools tuned for the biomedical domain.

Current developments focus on the use of learning methods in the extraction process. These methods are introduced at different levels in the system architecture. A first use is the acquisition of domain knowledge to enhance the extraction phase. A second use concerns a dynamic adaptation of existing modules during the analysis according to specific features in a text or to specific text genres.